\documentclass[twocolumn,runningheads]{llncs}
\usepackage{hyperref}
\usepackage{graphicx}
\usepackage{xcolor}
\usepackage[nice]{nicefrac}
\usepackage{amsmath}
\usepackage{amsfonts}
\usepackage{mathtools}
\usepackage{multirow}
\usepackage[textwidth=18.31cm,top=1.5cm]{geometry}
\usepackage{amsmath,amssymb,amsfonts}

\usepackage{algorithmic}
\usepackage{textcomp}
\usepackage{xcolor}
\usepackage{booktabs}
\usepackage{multirow}
\usepackage{algorithm}

\usepackage{url}
\def\BibTeX{{\rm B\kern-.05em{\sc i\kern-.025em b}\kern-.08em
    T\kern-.1667em\lower.7ex\hbox{E}\kern-.125emX}}

\graphicspath{{./images/}}  
\begin{document}
\title{Reducing Domain Gap with Diffusion-Based Domain Adaptation for Cell Counting}
%
\twocolumn[
  \begin{@twocolumnfalse}
\author{Mohammad Dehghanmanshadi\inst{1} \and
Wallapak Tavanapong\inst{1} }

\institute{Computer Science Department ,
Iowa State University, IA, USA
\\\email{mdehghan@iastate.edu, tavanapo@iastate.edu}}

\maketitle
\begin{abstract}
Generating realistic synthetic microscopy images is critical for training deep learning models in label-scarce environments, such as cell counting with many cells per image. However, traditional domain adaptation methods often struggle to bridge the domain gap when synthetic images lack the complex textures and visual patterns of real samples. In this work, we adapt the Inversion-Based Style Transfer (InST) framework originally designed for artistic style transfer to biomedical microscopy images. Our method combines latent-space Adaptive Instance Normalization with stochastic inversion in a diffusion model to transfer the style from real fluorescence microscopy images to synthetic ones, while weakly preserving content structure.

We evaluate the effectiveness of our InST-based synthetic dataset for downstream cell counting by pre-training and fine-tuning EfficientNet-B0 models on various data sources, including real data, hard-coded synthetic data, and the public Cell200-s dataset. Models trained with our InST-synthesized images achieve up to 37\% lower Mean Absolute Error (MAE) compared to models trained on hard-coded synthetic data, and a 52\% reduction in MAE compared to models trained on Cell200-s (from 53.70 to 25.95 MAE). Notably, our approach also outperforms models trained on real data alone (25.95 vs. 27.74 MAE). Further improvements are achieved when combining InST-synthesized data with lightweight domain adaptation techniques such as DACS with CutMix. These findings demonstrate that InST-based style transfer most effectively reduces the domain gap between synthetic and real microscopy data. 
Our approach offers a scalable path for enhancing cell counting performance while minimizing manual labeling effort. The source code and resources are publicly available at: \url{https://github.com/MohammadDehghan/InST-Microscopy}.

\end{abstract}
  \end{@twocolumnfalse}
]

\section{Introduction}
\label{sec:intro}
One of the major challenges in the biomedical imaging domain is the limited availability of labeled data for training deep learning models \cite{10984423,zhang2024mapseg}. Cell counting is essential for various biomedical research and clinical applications, including stem cell research and cancer diagnosis \cite{molnar2016accurate}. Automated cell counting faces several challenges. For instance, the number of cells per fluorescence microscopy image can range from a few to thousands, with many images containing crowded and overlapping cells \cite{mohammed2023idcia}. Manual labeling of individual cells is tedious, time-consuming, and requires significant domain expertise. Moreover, even the largest publicly available datasets—often providing only approximate cell locations—contain just a few hundred images \cite{mohammed2023idcia}. This scarcity of annotated data restricts the generalization and performance of deep learning models and motivates the development of quality synthetic data to supplement real training samples.

One promising direction is to generate synthetic microscopy images that are both structurally accurate and visually realistic. However, a persistent challenge in this Sim2Real (synthetic-to-real) paradigm is the domain gap between real and synthetic images, which often causes significant performance degradation when models trained on synthetic data are applied to real-world test sets. Recent advances in medical image synthesis include anatomy/pathology-guided generation for chest X-rays \cite{chen2024medical}, and physics-based 3D modeling for dermatology \cite{kim2024s}. While these approaches produce high-fidelity images, they do not always enable fine-grained control over spatial structure. On the other hand, hard-coded synthetic data allows precise control of cell count and layout, but lacks the realistic textures, noise, and color variations of real microscopy images. This visual domain gap substantially limits the transferability of models trained on synthetic data.

We introduce \textbf{InST-Microscopy}, an adaptation of the Inversion-Based Style Transfer (InST) framework \cite{zhang2023inversion} for generating realistic, weakly structure-preserving synthetic fluorescence microscopy images. Our approach integrates Adaptive Instance Normalization (AdaIN)-based latent initialization \cite{huang2017arbitrary} and stochastic inversion, as introduced in InST, and tailors them for the Sim2Real (synthetic-to-real) setting in fluorescence microscopy. Unlike previous style transfer or medical image synthesis methods, such as Generative Adversarial Networks (GANs) or text-guided diffusion models, our method transfers fine-grained visual style from real to synthetic images in a data-efficient manner, requiring only minimal supervision through lightweight training (via textual inversion). This makes InST-Microscopy particularly effective for structure-sensitive biomedical tasks such as cell counting.

We evaluated InST-Microscopy on a downstream cell counting task, demonstrating that supplementing real data with our style-transferred synthetic dataset improves model performance over training with real data alone. Furthermore, we show that DACS \cite{tranheden2021dacs}, a lightweight domain adaptation method, in combination with CutMix \cite{yun2019cutmix}, can further enhance generalization when used with our synthetic data.

In summary, our contributions are as follows.

\begin{itemize}
\item We propose InST-Microscopy, a new style transfer method tailored to Sim2Real in biomedical imaging, enabling realistic image synthesis with weak structural preservation and minimal supervision.
\item We conducted extensive quantitative and qualitative evaluations, demonstrating that our InST-synthesized synthetic data yields a further 13\% reduction in MAE over the best hard-coded synthetic baseline, a 52\% reduction over the Cell200-s baseline, and achieves improved performance relative to models trained solely on real data. 
\item We release the largest structure-aware synthetic microscopy dataset to date (600+ images with masks), along with code and pre-trained models to support reproducible Sim2Real research in microscopy.
\end{itemize}

\section{Related Work}
\subsection{Sim2Real in Medical Imaging}

Sim2Real transfer is crucial in biomedical imaging, particularly in tasks such as cell counting, where labeling many cells is time-consuming. Sim2Real could offer a scalable alternative. However, the domain gap between real and synthetic images often leads to significant performance degradation. To address this gap, researchers have explored various solutions, including domain adaptation~\cite{zhu2022fine,zhu2023daot}, single-domain generalization~\cite{peng2024single}, and generative modeling strategies such as knowledge diffusion~\cite{xie2023striking}.

In the context of cell detection, recent work has proposed diffusion-based layout generation methods that explicitly incorporate spatial patterns to better simulate biological plausibility~\cite{li2024spatial}. These approaches show that introducing structural priors during synthetic image generation can enhance downstream detection performance. Unlike these methods, our approach directly modifies image appearance using diffusion-based style transfer, while weakly preserving content structure. 

\subsection{Diffusion Models for Style Transfer}

Diffusion models have emerged as powerful generative tools, outperforming traditional GANs in various image synthesis tasks~\cite{rombach2022high}. Style transfer models, such as StyleFormer~\cite{wu2022styleformer} and StyTr2~\cite{deng2022stytr2}, have demonstrated the ability to model rich and complex visual patterns. However, these methods focus primarily on artistic style transfer and often lack the structural precision required in biomedical applications.

Recent advances have pushed diffusion models into applications demanding spatial and semantic consistency, such as crowd counting and density estimation. For example, Zhu et al.~\cite{zhu2022fine} utilize a fine-grained fragment diffusion approach to align distributions across domains, minimizing discrepancies between corresponding image regions for improved cross-domain crowd counting—without relying on adversarial objectives, while DAOT~\cite{zhu2023daot} leverages optimal transport to align domain-agnostic features. In contrast, we aim to directly transfer visual style at the image level from real microscopy images to synthetic ones, without relying on fine-grained annotations or adversarial alignment. 

Additionally, guidance-enhanced models such as ControlNet~\cite{zhang2023adding} introduce structural constraints using external cues (e.g., edge maps, segmentation masks). Although effective, such methods require dense annotations that are costly or infeasible in microscopy.

\subsection{Inversion-Based Style Transfer with Diffusion}

Inversion in diffusion models refers to mapping a real image back to its corresponding latent noise and conditioning vectors such that the model can regenerate the image through forward sampling. This inversion process enables example-guided generation, where the aim is to transfer visual attributes (e.g., style) while preserving semantic or spatial content. 

Textual inversion~\cite{gal2022image} and DreamBooth~\cite{ruiz2023dreambooth} are two prominent methods for incorporating new visual concepts into text-to-image diffusion models such as Stable Diffusion. Textual inversion learns pseudo-token embeddings from a few example images without modifying the model weights, while DreamBooth fine-tunes the model on a target concept using a small dataset. Although both methods are effective for integrating novel objects or styles, they do not preserve the spatial layout of the original image, limiting their applicability in domains like biomedical imaging where spatial fidelity is essential. 

Recently, Zhang et al.~\cite{zhang2023inversion} introduced InST (Inversion-based Style Transfer), a method that encodes the style of a single reference image as a pseudo-textual token and uses it to guide diffusion-based image generation. This approach offers precise style control while avoiding the complexity of textual prompt engineering. However, InST was originally developed for artistic domains, where spatial structure is less critical.

In contrast, we propose the first adaptation of InST to biomedical imaging. Our method transfers style from real fluorescence microscopy images to synthetic ones in the latent space using Adaptive Instance Normalization (AdaIN)  and stochastic inversion, while aiming to preserve the approximate spatial arrangement of cells. This enables the generation of realistic, structurally consistent synthetic images to support weakly supervised cell counting.

\section{Proposed Method}

Our objective is to generate realistic microscopy images from synthetic hard-coded content images, while preserving the original cell layout and count. This is critical for training robust cell counting models in settings where real annotated data is scarce or expensive to obtain. The hard-coded content images are synthetically generated using predefined rules that mimic cell distribution patterns, controlling attributes such as cell size, count, and spatial layout. However, these synthetic images lack the complex textures of real microscopy data, resulting in a visual domain gap that hinders generalization. 

\begin{figure*}[htbp]
\centering
\includegraphics[width=\textwidth]{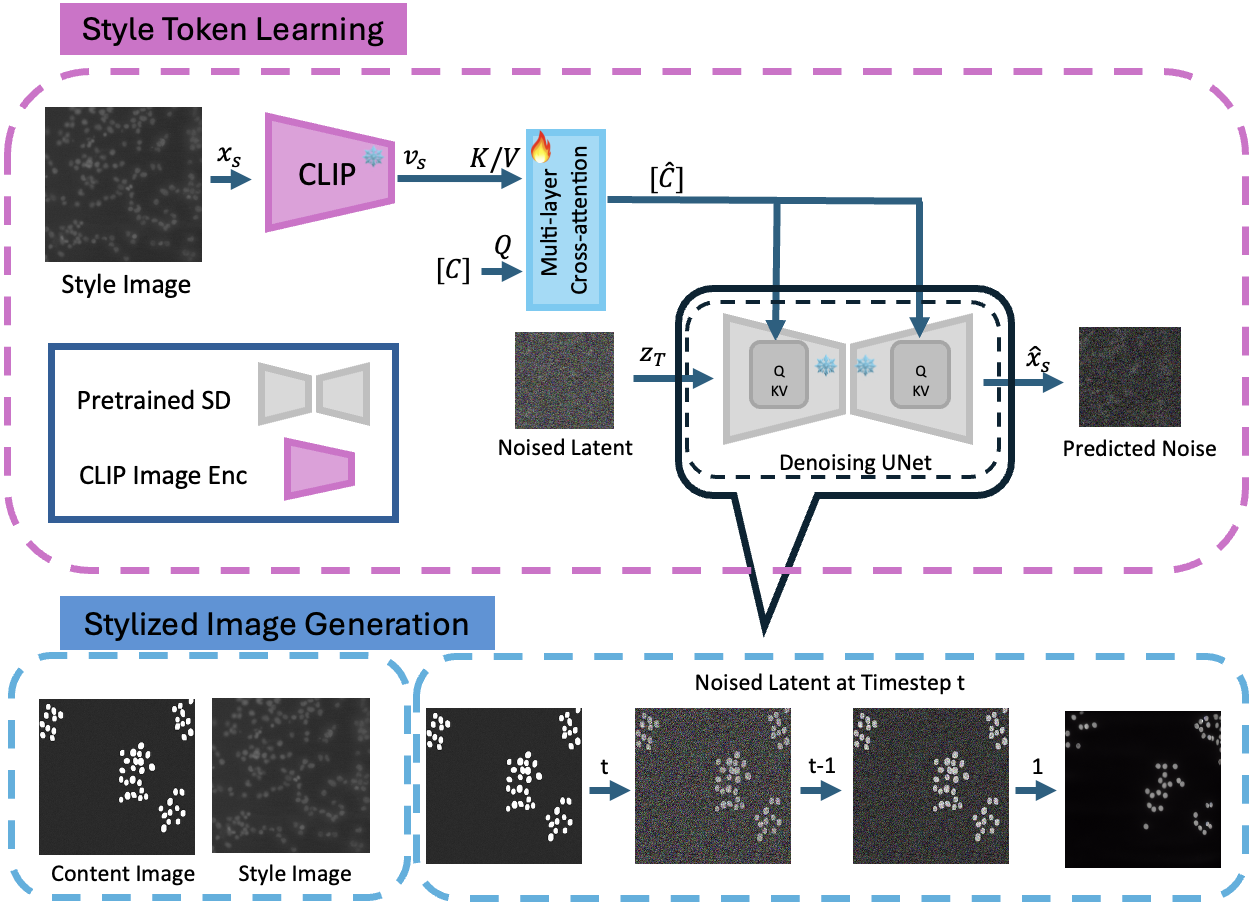}
\caption{
Overview of the proposed pipeline based on the Inversion-Based Style Transfer (InST) framework. 
\textbf{Top: Style Token Learning}. A CLIP image encoder extracts a visual embedding from a style image, which is passed through a multi-layer cross-attention module to learn a pseudo-token $[\hat{C}]$. This token is optimized by minimizing the denoising loss within the latent diffusion model (LDM) to reconstruct the original style image from noise. 
\textbf{Bottom: Stylized Image Generation}. During inference, content and style images are encoded into latents and combined using AdaIN to produce an initialized latent. A controlled amount of noise is added to this latent to obtain $z_t$, which is passed to the LDM and conditioned on the learned style token $[\hat{C}]$ to generate a stylized output image. 
Note: SD = Stable Diffusion; AdaIN is only used during inference; $[C]$ is the learnable style token, and $[\hat{C}]$ is its mapped version for conditioning.}
\label{fig:inst_pipeline}
\end{figure*}

To achieve this, we adapt the Inversion-Based Style Transfer (InST) framework to the microscopy domain, leveraging its stochastic inversion mechanism and incorporating latent initialization via AdaIN for improved style alignment. This section outlines the architecture, mathematical formulation, and each step of the proposed pipeline. Fig.~\ref{fig:inst_pipeline} illustrates the overall framework.

\subsection{Overview}

Let $x_c \in \mathbb{R}^{H \times W \times 3}$ denote the \textit{content image} (a hard-coded synthetic microscopy image), and $x_s \in \mathbb{R}^{H \times W \times 3}$ denote the \textit{style image} (a real microscopy image). The goal is to generate an output image $\hat{x} \in \mathbb{R}^{H \times W \times 3}$ that preserves the spatial structure of $x_c$ while adopting the appearance of $x_s$.

\subsection{Style Token Training via Textual Inversion}

To enable InST to replicate the style of a given image during generation, we must first learn a textual embedding that encapsulates this style. This is accomplished through a training phase that produces a style token, denoted $[C]$, which is later used during inference to condition the generation.

Given a real style image $x_s$, we aim to find a pseudo-token embedding $[C] \in \mathbb{R}^d$ such that the output image $\hat{x}$ generated by the diffusion model, when conditioned on $[C]$, resembles $x_s$ in appearance. We denote the learnable style token as $[C]$, which is optimized through a cross-attention mapping $\mathcal{F}_\theta$ to produce the conditioning token $[\hat{C}]$ used during generation.

\subsubsection{CLIP-Guided Initialization}

We use a pre-trained CLIP image encoder $\text{CLIP}_{\text{img}}(\cdot)$ to extract a style feature vector $v_s \in \mathbb{R}^d$:

\begin{equation}
v_s = \text{CLIP}_{\text{img}}(x_s).
\end{equation}

This vector captures high-level semantics of the style image. It is passed through a lightweight attention module $\mathcal{F}_{\theta}$ to produce a token embedding $[\hat{C}]$:

\begin{equation}
[\hat{C}] = \mathcal{F}_{\theta}(v_s),
\end{equation}

where $\mathcal{F}_{\theta}$ is implemented as a stack of cross-attention blocks followed by layer normalization. 
As proposed in InST, we replace gradient-based token optimization in textual inversion \cite{gal2022image} with a learned feedforward attention mechanism, which maps CLIP image embeddings to the CLIP text space, enabling faster convergence and improved style controllability.

\subsubsection{Diffusion-Guided Reconstruction Loss}

We initialize the diffusion model with noise $z_T \sim \mathcal{N}(0, I)$ where $I$ is the identity matrix of appropriate dimension and generate an image $\hat{x}_s$ using the diffusion process conditioned on $[\hat{C}]$:

\begin{equation}
\hat{x}_s = \text{StableDiffusion}(z_T \mid [\hat{C}]).
\end{equation}

To train the attention module $\mathcal{F}_{\theta}$ and produce a style embedding $[\hat{C}]$, we minimize the denoising loss from the latent diffusion model (LDM), comparing the predicted noise with the actual noise added during forward diffusion. This loss ensures that the style token $[\hat{C}]$ guides the model toward reconstructing the original style image $x_s$ when starting from noise (see ‘Denoising UNet’ in Fig. \ref{fig:inst_pipeline}, which corresponds to the LDM UNet).

\subsection{Stylized Image Generation}
\subsubsection{Latent Encoding}

Both the content and style images are first encoded into a latent space using the pre-trained Variational Autoencoder (VAE) encoder from the publicly available Stable Diffusion model~\cite{rombach2022high}. This VAE is not trained or fine-tuned as part of our method. Formally, let $\mathcal{E}(\cdot)$ denote the pre-trained encoder, and $\mathcal{Z}_c = \mathcal{E}(x_c)$, $\mathcal{Z}_s = \mathcal{E}(x_s)$ be the resulting latent representations of the content and style images, respectively, where $\mathcal{Z}_c, \mathcal{Z}_s \in \mathbb{R}^{C \times H' \times W'}$; $C$ is the number of latent channels; $H’, W’$ are $H/8$ and $W/8$, respectively.

\subsubsection{Latent Initialization via AdaIN}

To align the global statistics of the latent features, we apply AdaIN \cite{huang2017arbitrary}, which adjusts the content features to have the same mean and variance as the style features:

\begin{equation}
\text{AdaIN}(\mathcal{Z}_c, \mathcal{Z}_s) = \sigma(\mathcal{Z}_s) \left( \frac{\mathcal{Z}_c - \mu(\mathcal{Z}_c)}{\sigma(\mathcal{Z}_c)} \right) + \mu(\mathcal{Z}_s),
\end{equation}

where $\mu(\cdot)$ and $\sigma(\cdot)$ denote the channel-wise mean and standard deviation computed over spatial dimensions.

The result is an initialized latent representation $\mathcal{Z}_{init}$ that blends structural information from $x_c$ with the statistical characteristics of $x_s$:

\begin{equation}
\mathcal{Z}_{init} = \text{AdaIN}(\mathcal{Z}_c, \mathcal{Z}_s).
\end{equation}

\subsubsection{Style Conditioning via Textual Inversion}

The learned token $[\hat{C}]$ from the textual inversion phase is reused during inference to condition the generation. This replaces a natural language prompt and directly encodes the visual style of $x_s$.

\subsubsection{Stochastic Inversion}

Rather than starting from pure Gaussian noise, we apply stochastic inversion to encode $\mathcal{Z}_{init}$ to a noisy latent $\mathcal{Z}_t$ at timestep $t$:

\begin{equation}
\mathcal{Z}_t = \sqrt{\bar{\alpha}_t} \cdot \mathcal{Z}_{init} + \sqrt{1 - \bar{\alpha}_t} \cdot \epsilon, \quad \epsilon \sim \mathcal{N}(0, I),
\end{equation}

where $\bar{\alpha}_t$ is the cumulative product of noise schedule coefficients up to timestep $t$, and $\epsilon$ is sampled noise.

This process mimics the forward diffusion used in training, but initializes the model with a partially noised version of a meaningful content latent. The timestep $t$ is determined by a \textit{strength parameter} $\gamma \in [0, 1]$, typically set as $t = \gamma \cdot T$ for a total of $T$ diffusion steps. The value of $\gamma$ controls the trade-off between content preservation and style strength — lower values retain more of the original layout, while higher values inject more style and abstraction.

Finally, the stylized image $\hat{x}$ is generated by reversing the diffusion process starting from $\mathcal{Z}_t$, conditioned on the style token $[\hat{C}]$ and optionally an unconditional prompt. This is done using the DDIM sampling strategy \cite{song2021denoising}: 

\begin{equation}
\hat{\mathcal{Z}} = \text{DDIM\_decode}(\mathcal{Z}_t \mid [\hat{C}]), \quad \hat{x} = \mathcal{D}(\hat{\mathcal{Z}}),
\end{equation}

where $\mathcal{D}$ is the VAE decoder of the diffusion model. The output $\hat{x}$ contains the texture and color style of $x_s$ while roughly preserving the spatial layout of $x_c$.

In summary, our method combines:
\begin{itemize}
    \item AdaIN for latent initialization with global style statistics,
    \item Textual inversion for style conditioning via a learned token,
    \item Stochastic inversion for soft structure preservation during sampling.
\end{itemize}

This approach allows the generation of realistic images from controlled synthetic layouts, suitable for training models on tasks like cell counting.

\section{Experiments}

We conducted comprehensive experiments to evaluate the effectiveness of our proposed diffusion-based style transfer method in generating realistic microscopy images and improving downstream cell counting performance. We assess both visual quality and task-specific generalization under various training strategies.

\subsection{Datasets}

\textbf{Synthetic Hard-Coded (Syn-HC)}: We generated synthetic microscopy images by programmatically placing cell-like ellipses with controlled spatial distribution and count. We chose ellipses to mimic the typical shape variability of cell nuclei observed in DAPI-stained cell images. In our analysis of 50 representative images from the IDCIA dataset~\cite{mohammed2023idcia}, most of the cells appeared approximately elliptical, with moderate variation in size and orientation. Each synthetic image was initialized with Gaussian background noise and populated with cells grouped around 3 to 7 randomly selected cluster centers. Cell clustering reflects the natural tendency of cells to form colonies or localized groupings in tissue samples, as commonly observed in fluorescence microscopy images from the IDCIA dataset. The total number of cells per image was sampled from a normal distribution (mean $\mu = 120$, std $\sigma = 90$), clamped to a maximum of 600.  Each cell was rendered as a randomly shaped and rotated ellipse with axis lengths sampled around a predefined cell size (15 pixels), and intensity sampled from a uniform range ($[100, 200]$). This choice was motivated by the empirical distribution of cell counts in IDCIA. To avoid biologically implausible cell overlap while allowing for natural variability, cells were placed such that the fraction of their area overlapping with previously placed cells did not exceed a small threshold (10\%). Postprocessing includes random contrast and brightness adjustment. The full procedure is detailed in Algorithm~\ref{alg:synthetic_generation}. This process also produces precise binary ground truth masks marking cell centers, enabling their use in supervised training and evaluation. 

\textbf{Real Microscopy Images (Real)}: This set consists of all 119 DAPI-stained real fluorescence microscopy images from the IDCIA dataset.  Each image is annotated with the total cell count and the $(x, y)$ coordinates of individual cells. We follow the original split provided by the IDCIA dataset: 95 images for training and 24 for validation. Images from the \textit{Real} dataset were used as style images and for model performance evaluation. Note that the size of the data set in this domain is typically much smaller than that of generic images.

\begin{algorithm}[H]
\caption{Synthetic Microscopy Image Generation with Clustered Cells}
\label{alg:synthetic_generation}
\begin{algorithmic}[1]
\STATE \textbf{Input:} Image size $(H, W)$; mean $\mu$ and std $\sigma$ for cell count; cluster variance $\delta$; intensity range $[i_{\min}, i_{\max}]$
\STATE \textbf{Output:} Synthetic image $I$, ground truth mask $G$, cluster centers $\mathcal{C}$

\STATE Sample number of clusters $C \sim \mathcal{U}(3, 7)$
\STATE Sample total number of cells $n \sim \mathcal{N}(\mu, \sigma)$, clamp $n$ to $[1, 1500]$
\STATE Initialize $I$ with Gaussian noise; $G \leftarrow 0$
\STATE Randomly select $C$ cluster centers $\mathcal{C}$ in the image
\STATE Assign $n$ cells evenly to clusters
\STATE Initialize occupancy mask $M \leftarrow \text{False}^{H \times W}$
\FOR{each cluster center $(y_c, x_c) \in \mathcal{C}$}
    \STATE Set placed cell counter and attempt counter to zero
    \WHILE{(not all assigned cells placed) \textbf{and} (attempts $<$ max)}
        \STATE Sample ellipse axis lengths $(a, b)$, angle $\theta \sim \mathcal{U}[0, \pi]$
        \STATE Sample position $y \sim \mathcal{N}(y_c, \delta),\ x \sim \mathcal{N}(x_c, \delta)$
        \STATE Sample intensity $i \sim \mathcal{U}(i_{\min}, i_{\max})$
        \STATE Create ellipse at $(y, x)$ with axes $(a, b)$, angle $\theta$, intensity $i$
        \STATE Compute the fraction of the ellipse area overlapping with occupied regions in $M$
        \IF{overlap fraction $<$ 10\%} 
            \STATE Add ellipse to $I$ (max composite), mark area in $M$
            \STATE Set $G[y, x] \leftarrow 255$ \COMMENT{cell center only}
            \STATE Increment placed cell counter
        \ENDIF
    \ENDWHILE
\ENDFOR
\STATE Apply random global contrast/brightness: $I \leftarrow \alpha I + \beta$
\STATE Clip $I$ to $[0, 255]$ and convert to uint8
\RETURN $I$, $G$, $\mathcal{C}$
\end{algorithmic}
\end{algorithm}

\textbf{Cell200-s}: A subset of the publicly available Cell-200 dataset \cite{ding2022continuous}. The full dataset has  200,000 64x64 simulated fluorescence microscopy images with varying cell types, densities, and noise levels. It is used to test cross-dataset generalization. Cell200-s has 1,000 images selected from Cell-200 to ensure a balanced distribution across different bins of cell densities. Specifically, cell counts from Cell-200 were binned into four intervals ([0–50), [50–100), [100–150), [150–200)), and 250 image samples were randomly drawn from each bin to ensure representation across density levels. The selected subset was further divided into 800 images for training and 200 images for validation using a stratified approach, so that each cell count range was equally represented in both splits. 

\textbf{InST Outputs (Ours)}: The set of synthetic images stylized using our proposed framework by transferring the appearance from real DAPI-stained images onto synthetic ones while weakly preserving structural content. To assess performance across varying cell densities, we selected 12 synthetic images divided into three categories based on cell count: low ($<100$), medium ($100–200$), and high ($>200$), with 4 images randomly sampled per category. We chose 4 images per category to ensure a minimal yet representative sample, allowing for both visual diversity and consistent qualitative assessment across density levels. Fewer images (e.g., 1 or 2 per category) would risk insufficient coverage of the structural variability present within each density range.

For generating InST images, we used the official implementation provided by the original authors\footnote{\url{https://github.com/zyxElsa/InST}}, adapting it to our microscopy data and style strength settings. For each cell density category (low, medium, high), we trained a single pseudo-token embedding using textual inversion. Each pseudo-token was optimized over 1,000 training steps with a learning rate of $5 \times 10^{-4}$, taking approximately 1.5 hours per token (batch size 1, $<10$ GB GPU memory). Following the original InST framework, the learned style tokens were incorporated into the multi-layer cross-attention diffusion model with 8 attention heads, a context dimension of 768, and a style strength parameter of 0.7 (selected via pilot experiments to balance structure and style).

Table~\ref{tab:dataset-stats} summarizes the key statistics of the four datasets. 

\begin{table}[htbp]
\centering
\caption{Descriptive statistics of datasets}
\label{tab:dataset-stats}
\begin{tabular}{lccc}
\toprule
Datasets      & \#Images & Mean $\pm$ Std. \#Cells   & Image Size \\
\midrule
~~ Real         & 119      & 141.35 $\pm$ 122.56                   & 600 $\times$ 800 \\
~~ Syn-HC/InST-generated & 618      & 101.32 $\pm$ 76.43                   & 256 $\times$ 256 \\
~~ Cell200-s      & 1000      & 100.33 $\pm$ 57.28               & 64 $\times$ 64 \\
\bottomrule
\end{tabular}
\end{table}

\subsection{Evaluation Metrics}

We adopt the following commonly used metrics for counting tasks. Low MAE and RMSE are desirable.
\begin{itemize}
    \item \textbf{Mean Absolute Error (MAE)}: Measures average absolute deviation between predicted and ground truth cell counts.
    \item \textbf{Root Mean Squared Error (RMSE)}: Measures the square root of the average squared differences between predicted and ground truth counts, penalizing larger errors more heavily.
\end{itemize}

\subsection{Training and Testing of Cell Counting Models}

We used EfficientNet-B0 \cite{tan2019rethinking} as the backbone for all cell counting models. 
 We compare the following training regimes:

\begin{itemize}
    \item \textbf{Fine-tuned}:  The model was first trained on one synthetic dataset (Syn-HC, InST-generated, or Cell200-s), then all layers were fine-tuned end-to-end on the real IDCIA training set (95 images). This approach leverages transfer learning, using synthetic data to pretrain before adaptation to real images.
    \item \textbf{Dual Training}: The model was trained jointly on batches from real and synthetic data. Each mini-batch contains equal numbers of images from both domains (8 real + 8 synthetic per batch). Dual training was conducted for Syn-HC, InST-generated, and Cell200-s datasets.
    \item \textbf{DACS+CutMix}: 
    We adapted the official DACS implementation\footnote{\url{https://github.com/vikolss/DACS}} and replaced ClassMix with CutMix, which is more appropriate for regression-based tasks like cell counting. For CutMix, we used $\alpha = 1.0$ for the Beta distribution and applied mixing with a probability of 0.5. Hyperparameters were selected based on a small grid search over $\alpha \in \{0.5, 1.0, 2.0\}$ using validation MAE, evaluated on the held-out 24-image real validation set from IDCIA. In all DACS experiments, we computed the ground truth cell count for each CutMix-augmented image by creating cell location maps for both real and synthetic sources, applying CutMix to these maps alongside the images, and then counting the resulting cell centers in the mixed mask. This ensures accurate regression targets for each composite image.
\end{itemize}

The real IDCIA images are $600 \times 800$ pixels, while Syn-HC and InST images are $256 \times 256$, and Cell200-s images are $64 \times 64$. To handle this resolution mismatch, we used the following setup:

\textbf{Training:}  
For Syn-HC and InST, models were trained on random $200 \times 200$ crops (one per image per epoch). For Cell200-s, full $64 \times 64$ images were used.

\textbf{Evaluation on Real Images:}  
We applied a non-overlapping sliding-window approach. Syn-HC/InST-trained models used $200 \times 200$ windows; Cell200-s-trained models used $64 \times 64$ windows. Final predictions were obtained by aggregating window-level outputs.

Training was performed using the Adam optimizer with a learning rate of $1e{-4}$, weight decay of $1 \times 10^{-5}$, batch size 16 (8 per domain for dual training) for 50 epochs. All models were initialized with ImageNet pre-trained weights. 

\section{Results}

Table~\ref{tab:all_results_suggested} shows that using our InST-syn dataset is better at closing the domain gap than Cell200-s or our hard-coded synthetic dataset, regardless of the training strategies. With a fine-tuning strategy, our InST-syn dataset brings the MAE to a lower value than that of the supervised training on Real. Given a particular dataset, DACS+CutMix was the best training strategy among all three to narrow the domain gap.

\subsection{Fine-tuning on Synthetic Data}

When models are pre-trained on synthetic data and fine-tuned on a small set of real images, the choice of synthetic source strongly affects domain gap. As shown in Table~\ref{tab:all_results_suggested}, Cell200-s and Syn-HC baselines have large gaps to real-data-only training ($\Delta$MAE = 27.86 and 13.46, respectively), while InST-Syn not only closes this gap but actually improves over the real-data-only baseline ($\Delta$MAE = –1.49). This demonstrates that our InST-style transfer effectively aligns synthetic and real data distributions, making synthetic pretraining beneficial for downstream cell counting.

The improved performance of Syn-HC over Cell200-s may be attributed to its larger image size and/or more realistic visual characteristics, such as closer cell morphology or intensity distributions to real data. Further analysis would be valuable to disentangle these effects.

\subsection{Dual Training with Real and Synthetic Data}
In dual training, each mini-batch includes both real and synthetic images. This joint exposure further reduces the domain gap for all synthetic datasets, with InST-Syn again performing best ($\Delta$MAE = 0.04). This indicates that supplementing real data with InST-synthesized images provides additional benefit, bringing performance nearly indistinguishable from real-data-only training.

\subsection{Domain Adaptation with DACS + CutMix}

Applying DACS + CutMix domain adaptation yields the best overall results, particularly for InST-Syn, which achieves the lowest MAE (25.95) and the smallest gap to real-only training ($\Delta$MAE = –1.79). This confirms that the combination of our synthetic images and lightweight domain adaptation effectively closes, and in some cases reverses, the domain gap.

\begin{table}[t]
\centering
\caption{
Performance on the \emph{Real} validation set per combination of training strategy and dataset.
$\Delta$ indicates the difference in MAE/RMSE compared to supervised training on real data
(lower is better; negative indicates improvement).
}
\label{tab:all_results_suggested}

\resizebox{\linewidth}{!}{%
\begin{tabular}{llll}
\toprule
\textbf{Training Strategies} & \textbf{Dataset} & \textbf{MAE} ($\Delta$) & \textbf{RMSE} ($\Delta$) \\
\midrule
Supervised & \emph{Real} & \textbf{27.74} & \textbf{42.50} \\
\midrule
\multicolumn{4}{l}{\emph{Fine-tuned}} \\
& Cell200-s & 55.60 (27.86) & 73.10 (30.6) \\
& Syn-HC & 41.20 (13.46) & 61.45 (18.95) \\
& InST-Syn & \textbf{26.25 (-1.49)} & \textbf{42.12 (-0.38)} \\
\midrule
\multicolumn{4}{l}{\emph{Dual training}} \\
& Real+Cell200-s & 54.40 (26.66) & 59.92 (17.42) \\
& Real+Syn-HC & 33.10 (5.36) & 52.78 (10.28) \\
& Real+InST-Syn & \textbf{27.70 (0.04)} & \textbf{43.91 (1.41)} \\
\midrule
\multicolumn{4}{l}{\emph{DACS+CutMix} \cite{tranheden2021dacs}} \\
& Cell200-s & 53.70 (25.96) & 55.35 (12.85) \\
& Syn-HC & 29.85 (2.11) & 46.33 (3.83) \\
& InST-Syn & \textbf{25.95 (-1.79)} & \textbf{41.76 (-0.74)} \\
\bottomrule
\end{tabular}%
}
\end{table}

\subsection{Qualitative Results}

\begin{figure}[htbp]
\centering
\includegraphics[width=0.48\textwidth]{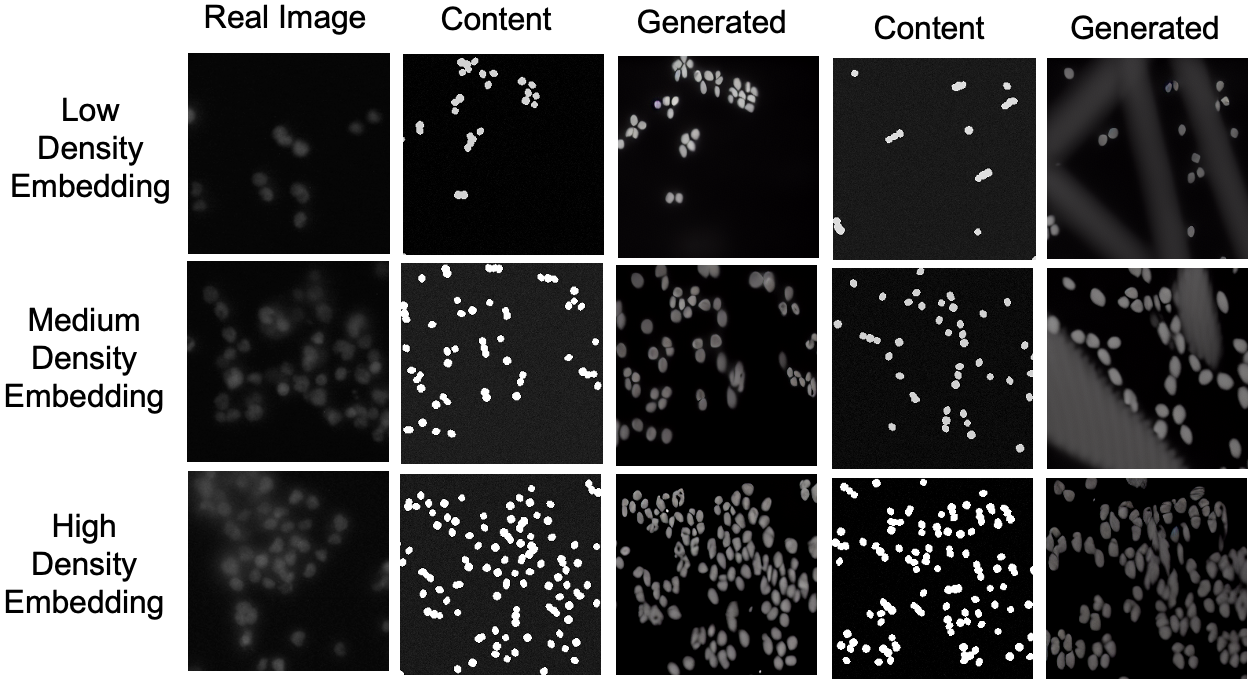}
\caption{
\textbf{Qualitative results for category-aware style transfer.} 
Each row corresponds to a different cell density category—low, medium, and high—based on the learned density-specific style token (embedding). The first column in each row shows a real DAPI-stained microscopy image representing the target visual style for that category. The following columns display pairs of synthetic content images (pre-stylization) and their corresponding generated outputs after style transfer with the appropriate density embedding.
}
    \label{fig:fig_2}
\end{figure}

We present qualitative results to evaluate the visual quality and structural consistency of the generated microscopy images. These examples demonstrate the effectiveness of our method in bridging the domain gap.
Figure~\ref{fig:fig_2} illustrates the effect of our method for three distinct cell density regimes: low [$0$–$100$), medium [$100$–$200$), and high ($>=200$) cell counts. 

In each row, the leftmost column displays a real DAPI-stained microscopy image representing the target style and cell density for that category. The remaining columns show two pairs of synthetic content images and their corresponding generated outputs, where each synthetic image is stylized using the density-specific learned style token. The generated images preserve the cell layout of the synthetic input while successfully transferring the texture, illumination, and noise characteristics of the real microscopy image. 

Notably, in the second row, third column (medium density), the generated image shows a blurred region with faint linear patterns. Such artifacts may result from similar features—like uneven illumination—in the real style images used during training. This illustrates that while our method effectively transfers realistic visual characteristics, it can also reproduce or amplify artifacts present in the style reference.

These examples demonstrate that our approach is able to generate visually realistic and structurally consistent images across a broad range of cell densities, effectively bridging the domain gap between synthetic and real microscopy data.

\begin{figure}[htbp]
\centering
\includegraphics[width=0.45\textwidth]{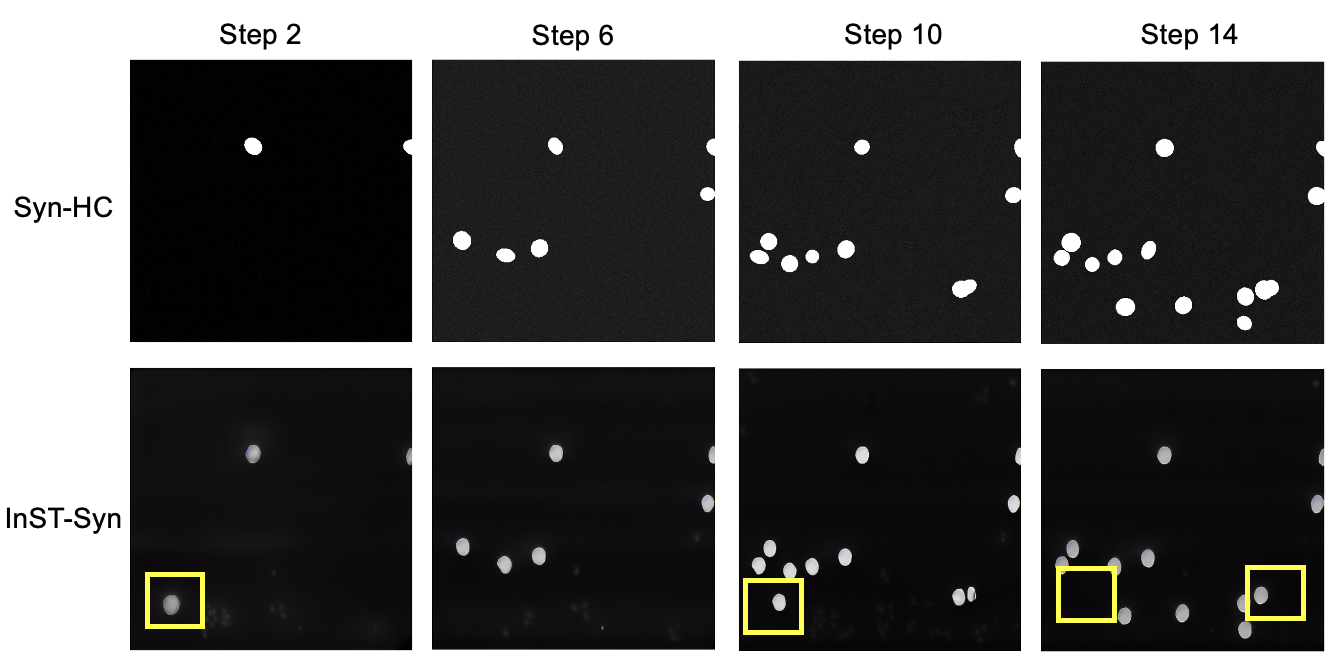}
\caption{
Qualitative analysis of structure preservation as the number of cells increases. Yellow boxes highlight cases of structural inconsistency in the generated images: \textbf{Step 2} — hallucination of an extra cell, \textbf{Step 6} — successful correction to match the intended layout, \textbf{Step 10} — another instance of hallucinated cell, and \textbf{Step 14} — merging of close cells (right box) and missing a cell (left box). 
}
    \label{fig:fig_3}
\end{figure}

Figure~\ref{fig:fig_3} highlights cases where structure preservation becomes more challenging as cell density increases. In this example, we gradually increase the number of cells in the input (Syn-HC, top row) one cell per step, and observe the stylized output (InST-Syn, bottom row). While the model generally preserves structure, it occasionally hallucinates or omits cells. For instance, in Step 2 and Step 10, an additional cell appears in the stylized image, whereas in Step 14, the model merges closely placed cells or fails to represent one of them. These examples reveal a limitation of the method in accurately preserving structure in densely populated regions, particularly when cell boundaries are ambiguous or overlapped. 

Together, these results demonstrate the visual realism and controllability of our style transfer framework, while also motivating future work on enforcing stricter structure preservation under complex scenarios.

\subsection{Efficiency and Resource Usage}

We evaluated the computational efficiency of our style transfer pipeline on a single NVIDIA Tesla V100 GPU (32GB VRAM). During inference, the generation of a $256 \times 256$ image using 50 DDIM sampling steps required approximately 10.1 seconds, with an average decoding time of 4.4 seconds per image. The inference process consumed a maximum of 7.68 GB of GPU memory (batch size 1, style strength 0.7).

 Since the diffusion backbone and CLIP encoder remained frozen during this phase, the training remained lightweight and efficient. Overall, the proposed method is computationally practical for research settings and scalable to multiple categories without the need for full model fine-tuning.

\subsection{Discussion and Limitations}

While our approach reduces the domain gap and improves cell counting accuracy, several limitations remain. We selected 618 out of 1,000 generated images, often excluding those with high cell density where spatial structure was poorly preserved.

\begin{figure}[htbp]
\centering
\includegraphics[width=0.45\textwidth]{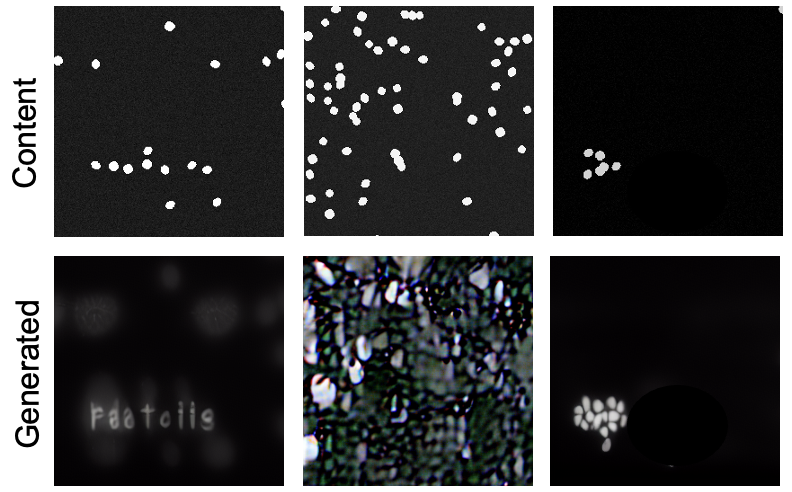}
\caption{
Examples of failure cases from our style transfer method. Each column illustrates a distinct failure mode: (left) non-cell-like or text-like artifacts, (middle) over-amplified background texture or noise, and (right) incomplete or partial style transfer.
}
    \label{fig:fig_4}
\end{figure}

Figure~\ref{fig:fig_4} shows representative failure cases, grouped by common artifact type: (1) non-cell-like or text-like artifacts, (2) over-amplified background texture/noise, and (3) incomplete style transfer or masking. These issues reflect challenges in preserving biological realism and spatial structure, especially in complex or noisy images.

Addressing these failure modes remains an important direction for future work.

\textbf{Ethical Considerations.} The use of synthetic images that visually resemble real microscopy data introduces ethical considerations, particularly regarding biological accuracy. Our model does not distinguish between live and dead cells, whereas in practice, expert biologists typically exclude dead cells from their counts. This discrepancy could lead to misinterpretation if such synthetic data is used without clear annotation or domain expertise. To mitigate this, it is crucial to transparently disclose the synthetic origin and limitations of such data in any downstream analysis or deployment.

\section{Conclusion}

We proposed a diffusion-based style transfer framework for generating realistic, structure-aware synthetic microscopy images to support scalable data generation in cell counting tasks. By leveraging Inversion-Based Style Transfer and Adaptive Instance Normalization within the diffusion model, our approach achieves improved realism and competitive downstream performance compared to other synthetic data baselines. Although our results highlight the promise of this method for cost-effective training data generation, challenges remain in faithfully preserving fine structural details, especially in high-density cases. Future work will focus on enhancing structure preservation, broadening expert validation, and extending the approach to additional biomedical imaging tasks.

\section*{Acknowledgment}
This work is partially supported by the Lange Faculty Excellence award and the National Science Foundation under Grant No. 2152117. 
Any opinions, findings, and conclusions or recommendations expressed in this material 
are those of the author(s) and do not necessarily reflect the views of the National Science Foundation.

\bibliographystyle{ieeetr}
\bibliography{refs_2}

\end{document}